\documentstyle{nemlap}

\newcommand{\argmax}[2]{\mbox{$\makebox[0mm]{\hspace{8ex}\raisebox{-1.5ex}{\tiny
$#1$}}{\rm argmax}\,[#2]$}}
\newcommand{\error}[1]{\mbox{\makebox[0mm]{\hspace{1.3ex}\raisebox{1.0ex}{\tiny
*}}#1}}  \newcommand{\prob}[1]{\mbox{${\sl
P}(#1)$}} \newcommand{\inputitem}[1]{\tt \refstepcounter{#1}
\raggedright \item \makebox[0mm]{\hspace{180ex} {\rm (\arabic{#1})}}}

\input{psfig}

\begin{document}

\title{Connected Text Recognition Using Layered HMMs and Token
Passing}

\author{Peter Ingels} \institute{Department of Computer and
Information Science \\ Link\"{o}ping University \\ S-581 83
Link\"{o}ping, Sweden\\ {\tt petin@ida.liu.se} }

\maketitle

\begin{abstract}
We present a novel approach to lexical error recovery on textual
input. An advanced robust tokenizer has been implemented that can not
only correct spelling mistakes, but also recover from segmentation
errors. Apart from the orthographic considerations taken, the
tokenizer also makes use of linguistic expectations extracted from a
training corpus. The idea is to arrange Hidden Markov Models (HMM) in
multiple layers where the HMMs in each layer are responsible for
different aspects of the processing of the input. We report on
experimental evaluations with alternative probabilistic language
models to guide the lexical error recovery process.
\end{abstract}

\bibliographystyle{plain}

\section{Introduction}
\label{sec:intro}
This paper presents the {\em layered Hidden Markov Model} (HMM)
technique set in the {\em Token Passing} (TP) framework
\cite{young+al:89}. The idea of having layers of HMMs encoding
different knowledge sources is well handled in the TP framework. The
TP model, which originates in speech processing, is independent of the
pattern matching algorithms used in for example {\em Connected Speech
Recognition} (CSR). The abstract model can be adopted to different
recognition algorithms and the recognition is viewed as a process of
passing tokens around a transition network. Furthermore it is easy to
interface and couple dependencies between the different knowledge
sources. The lexical errors that occur during text production are
misspellings and segmentation errors.  These can be further
categorized as real-word or nonword errors, and single or multiple
errors\footnote{A single error misspelling contains one instance of
one of the error types: character insertion, deletion, substitution
and transposition. A real-word error has occurred when a valid
(correctly spelled) word is substituted for another.}
(cf. Kukich~\cite{kukich:92b}). Since we are attempting to deal with
segmentation errors and not only regular spelling errors we adopt in
many ways the same view on the problem of tokenization as that of CSR
and we call it {\em Connected Text Recognition} (CTR). The idea is to
have a set of word modeling HMMs, the orthographic decoder, where the
individual HMMs assign a probability to a portion of the input symbol
stream being the word modeled by the HMM\@. Guiding the orthographic
decoder is the linguistic decoder. The linguistic decoder is also an
HMM (or several HMMs) that will assign probabilities to sequences of
words.

Although much effort has gone into the problem of spelling error
correction over the years, far less attention has been paid to the
closely related problem of correcting segmentation errors. One of the
few to address this problem is Carter~\cite{carter:92}. Carter
integrated an advanced tokenizer with the {\sc clare} system that
considers both spelling errors and segmentation errors when unknown
tokens are found in the input. The author does not describe the
(non-probabilistic) recovery methods in detail, but rather stresses
the need for syntactic and semantic knowledge to choose among the
multiple alternatives that may be hypothesized in the recovery
process. Carter's correction module could unambiguously correct 59 out
of 108 nonword error tokens in artificially generated sentences
without the use of domain-specific or contextual knowledge. For the
remaining 49 errors there were 224 correction hypotheses, including
all the correct ones. After syntactic and semantic knowledge had been
applied to disambiguate, 71 hypotheses remained and 5 of the correct
candidates had been eliminated.

In recent years interest has been directed towards probabilistic
methods in automatic spelling error detection/correction and
particularly the use of these methods to achieve context-sensitive
error recovery. Such methods need both orthographic knowledge, a
`noise-model' of some sort that almost always exploits a vocabulary,
and a model of word order. In most cases, however, researchers in the
area tend to emphasize one of the knowledge sources at the expense of
the other, thus limiting the scope of their techniques. Kernighan,
Church and Gale~\cite{kernighan+al:90,church+gale:91a} concentrate on
the `noisy channel' in their program {\sc correct} that can handle
single error nonword misspellings. Atwell and
Elliott~\cite{atwell+elliott:87} used the {\sc claws} part-of-speech
bigram language model to detect real-word errors and Mays {\em et
al.}~\cite{mays+al:91} used the trigram language model employed in the
IBM speech recognition project~\cite{bahl+al:83} to correct single
error real-word errors.

The technique presented here takes a more balanced approach to the
problem of lexical error recovery. The robust tokenizer at least has
the potential\footnote{The ability to deal with real-word errors
depends on the predictive power of the language model. Shortage of
data forces us to use a rather weak language model with which
real-word errors are hard to come to terms with.}  to handle all the
error types mentioned above. The robust tokenization process can be
viewed as the process of normalizing the character input stream
according to the vocabulary (orthographic decoder) and the language
model (linguistic decoder). Since the space character is just another
character, the segmentation error is merely the special case of
misspellings that involves the space character.

The following section gives an overview of the layered HMM technique
in the token passing framework. The experimental domain and
configurations along with the obtained results are presented in
section~\ref{sec:exp}~``Experiments''. The paper ends with
``Concluding~Remarks'' (section~\ref{sec:rem}).

\section{Layered HMMs and Token Passing}
\label{sec:hmm}
Although there can in general be multiple layers in the layered HMM
architecture we will focus on the two-layer setup, a single
utterance-modeling HMM in the topmost layer, the {\em Linguistic
Decoder} (LD), and a set of word modeling HMMs in the bottom layer,
the {\em Orthographic Decoder} (OD).

The problem of choosing the word $w_{i}$ out of a vocabulary $W =
w_{1}, w_{2},\ldots w_{M}$ that best matches a character sequence $C =
c_{1}, c_{2},\ldots c_{T}$ where $C$ is known to be a single word is
called the {\em Isolated Word Recognition}\footnote{Some form of IWR
is usually what is done in spell-checkers that come with commercial
word processors. Informal tests performed with {\sf iwr}, an
IWR-implementation of our approach, suggest that the technique
described here outperforms commercial spell-checkers by a good
margin.} problem (IWR). The question is then which word has the
highest probability given the character sequence, i.e. which word
$w_{i}$ maximizes $\prob{w_{i}|C}$. Bayes' rule states
that\label{p:bayes}
        \[\prob{w_{i}|C}=\frac{\prob{C|w_{i}} \prob{w_{i}}}{\prob{C}}\]
Choosing the word that best matches the character sequence is not
dependent on the probability of the sequence, so finding the $w_{i}$
that maximizes the numerator in Bayes' rule seems like a good
idea. The OD identifies each word $w_{i}$ in the vocabulary with an
HMM ${\cal M}_{w_{i}}$. The OD thus contains $M$ HMMs and each HMM
models one particular word form. The word that best matches the
character sequence is the one identified with ${\cal M}_{w_{i}}$ where
$\prob{C|{\cal M}_{w_{i}}}$ has the highest probability of all
HMMs. Looking at Bayes' rule, this number is the first factor in the
numerator. Making the obviously faulty (and soon to be revised)
assumption that all words are equiprobable, finding the word
$w_{i}^{*}$ that best matches the character sequence $C$ is simply
\[w_{i}^{*} = \argmax{W}{\prob{C|{\cal M}_{w_{i}}}}\]

\begin{figure}[h]
        \centerline{\psfig{figure=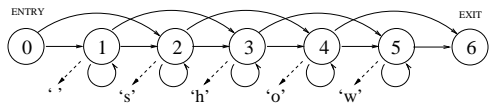}}
        \caption{The structure of ${\cal M}_{show}$\label{fig:m-show-space}}
\end{figure}

Figure~\ref{fig:m-show-space} shows the left-to-right OD HMM modeling
the word `show'. The states correspond to character positions in the
word modeled. The solid arrows represent transitions with non-zero
probabilities. The dashed arrows indicate what this particular model
is biased towards. State 2 for example can have non-zero probabilities
for all observables, but is strongly biased towards {\tt `s'},
i.e. $b_{2}(c)$ has the highest probability for $c=$\,{\tt `s'}. The
standard notation for HMM parameters is used here. The matrix
$\mbox{\boldmath $A$}$ ($a_{ij} = \prob{j|i}$) holds the state
transition distribution and the matrix $\mbox{\boldmath $B$}$
(\(b_{j}(v_{k}) = \prob{v_{k}|j}\)) holds the observation symbol
distribution, where $i$ and $j$ are states of the model and $v_{k}$ is
a symbol from the model's vocabulary/alphabet.

\noindent The structure of the word model in Figure~\ref{fig:m-show-space} is
slightly different from the standard Moore style HMM. The difference
is the two non-emitting states marked `entry' and `exit'. The entry
state is nothing more than the initial state distribution vector. The
exit state on the other hand adds the notion of final states to the
HMM\@. The final states of the model in Figure~\ref{fig:m-show-space}
are the ones connected to the absorbing exit state. The transitions
$a_{46}$ and $a_{56}$ determines with what probabilities state $4$ and
$5$ respectively are final states. The Baum-Welch reestimation
algorithm has to be slightly adjusted to account for the exit state
feature.

To perform isolated word recognition and connected text recognition,
the Viterbi algorithm is adopted to the token passing framework (see
also Young {\em et al.}~\cite{young+al:89}). A (partial) path, as
computed by the Viterbi algorithm, represents an alignment of states
in an HMM with the input characters. In the token passing algorithm
the head of such a path is represented by a token. A token contains
the {\em cost} of the path. The cost of a path is the negative
logarithm of the probability of the path. Thinking of an HMM as a
network of states, each state can hold one token\footnote{The number
of tokens left in a state after `the rest has been discarded' (see the
dashed box) actually determines the number alternative state sequences
that can be maintained, i.e. $n$ tokens per state implements $n$-best
recognition. For simplicity the pseudo-code describes 1-best
recognition.}. Extending the path forward in time (processing an input
character) means passing a copy of a states token to its connecting
states.

The notion of {\em time} inherent in the Viterbi algorithm refers to
reading characters from the input character stream. $c_{1}$ is read at
time $t=1$, $c_{2}$ is read at $t=2$ and so on. Since the HMMs of the
OD have characters as their observation symbols they are of course
time synchronized. The LD however, having words or OD HMMs as
observation symbols, is not time synchronous. The synchronous and
asynchronous variants are obviously different and this fact is
reflected in the token passing variants of the Viterbi algorithm.

The algorithmic outline below performs IWR. It is the TP variant of
the Viterbi algorithm applied to the synchronous OD HMMs. The portion
of the pseudo-code inside the dashed box is the {\em Step Model
Procedure} that will be reused in CTR.

\begin{tabbing}
Belo\=w:\\ \> The HMM has $N+1$ states numbered $0$ to $N$.\\ \> The
null token has cost $-\log 0 = \infty$.\\ \> The start token has cost
$-\log 1 = 0$
\end{tabbing}

\noindent {\bf Isolated Word Recognition with Token Passing}
\begin{tabbing}
At time \= $t=0$\\ \> put start token in the entry state\\ \> put null
tokens in all other states\\ {\bf for} \= {\bf each} time $t=1$ to $T$
{\bf do}\\ \dashbox{7}(330,120)[l]{ \> {\bf for} \= {\bf each} state
$i<N$ {\bf do}\\ \> \> Pass a copy of the token $Q$ in state $i$ to
all connecting states $j$\\ \> \> $Q.cost := Q.cost+(-\log
a_{ij})+(-\log b_{j}(c_{t}))$\\ \> discard all original tokens\\ \>
{\bf for} \= {\bf each} state $i<N$ {\bf do}\\ \> \> Find token with
$\min cost$ and discard the rest\\ \> {\bf for} \= {\bf each} state
$i$ connected to state $N$ {\bf do}\\ \> \> Pass a copy of the token
$Q$ in state $i$ to state $N$\\ \> \> $Q.cost := Q.cost+(-\log
a_{iN})$\\ \> Find token in state $N$ with $\min cost$ and discard the
rest\\ } \\ \> Put null token in entry state\\ $\Box$
\end{tabbing}

At time $T$ the isolated word recognizer inspects the exit state of
all the HMMs of the vocabulary and the model with the lowest cost in
its exit state is the one that best matches the character sequence.

In connected text recognition the character sequence \(C =
c_{1},c_{2},\ldots\,c_{T}\) can contain any number of misspelled and
ill-segmented words. The recognition task in CTR is to find the
correct set of HMMs and the alignment of them that best matches the
character sequence $C$.

State 1 in Figure~\ref{fig:m-show-space} is biased towards the space
character. To be able to deal with segmentation errors in CTR it is
crucial that inter-word space characters be modeled in some way. Note
that ${\cal M}_{show}$ in Figure~\ref{fig:m-show-space} will score a
maximum probability for the character sequence {\tt
`\symbol{32}show'}. An alternative approach would be to have the space
character be a word on its own, i.e. have an HMM that models gaps
between words. This is not such a good idea however since language
modeling would get unjustifiably expensive. Note that a word is just a
character sequence that has an HMM in the OD modeling it. Since the
space character is just another character it is quite alright to have
for example {\tt ` Winston Churchill'} or {\tt ` as soon as possible'}
be a word.

The job of the LD is to supply the pattern matching OD with
context. The context supplied by the LD is used to limit the search
space, enable real-word error correction and to decide on `close
calls'. For example, what is the correct repair for {\tt `\ldots in
the aboue table'}? Should {\tt `aboue'} be {\tt `above'} or {\tt
`about'}?

In our case the LD is a single HMM. The observables of the LD HMM are
the words of the vocabulary, or in other words, the observables of the
LD HMM are the word modeling HMMs of the OD\@. This is the trick of
the layered HMM approach, to find an explicit connection between the
LD and the OD\@. The LD schematically:
\begin{eqnarray}
\prob{w_{1},\ldots ,w_{T}} = \sum_{contexts}\
\prod_{i=1}^{T}\prob{w_{i}|context_{i}}\prob{context_{i}}\label{eq:schematic}
\end{eqnarray}
This refers back to the discussion on Bayes' rule above. The second
factor of the numerator of Bayes' rule that was assumed irrelevant is
now supplied by the LD\@.

The Token Passing algorithm is now set to recognize utterances instead
of isolated words. The LD HMM used in the superficial algorithmic
presentation below is of the same type as the OD HMM in
Figure~\ref{fig:m-show-space} except that it is not limited to
left-to-right transitions. Tokens are passed within an OD HMM
according to the topology of the model and the forwarding of tokens
from the exit state of one OD HMM to the entry state of another is the
job of the LD\@. The step model procedure for the IWR case is reused
here with only minor changes. The token put in the entry state of the
OD HMM does not have zero cost since it has been subject to prior cost
accumulation.

\begin{tabbing}
Belo\=w:\\ \> An OD HMM is {\em activated} when a non-null token is
put in its entry state.\\ \> An OD HMM is {\em deactivated} when all
states are assigned null tokens.
\end{tabbing}

\noindent {\bf Connected Text Recognition with Token Passing}
\begin{tabbing}
At \= time $t=0$\\ \> LD:\ \= Put start token in the entry state\\ \>
\> Put null tokens in all other states\\ \> OD:\> Deactivate all
models\\\\ {\bf For} \= {\bf each} time $t=1$ to $T$ {\bf do}\\ \>
LD:\ \= {\bf for} \= {\bf each} state $i<N$ with a non-null token {\bf
do}\\ \> \> \> Pass a copy of the token $Q$ in state $i$ to the entry
state of all\\ \> \> \> OD HMMs ${\cal M}_{w_{k}}$ that are observable
in state $j$ (${\cal M}_{w_{k}}$ are activated)\\ \> \> \> $Q.cost :=
Q.cost + (-\log a_{ij})+(-\log b_{j}(w_{k}))$\\ \> \> Put null tokens
in all states\\ \> OD:\ \= {\bf Step Model Procedure} with $c_{t}$\\
\> \> {\bf for} \= {\bf each} OD HMM with a non-null token in the exit
state {\bf do}\\ \> \> \> Propagate the token up to the LD state it
once came from\\ \> LD:\ \= {\bf for} \= {\bf each} state $i<N$ {\bf
do}\\ \> \> \> Find the token with $\min cost$ and discard the rest\\
\> \> {\bf for} \= {\bf each} state $i<N$ connected to state $N$ {\bf
do}\\ \> \> \> Pass a copy of the token $Q$ in state $i$ to state
$N$\\ \> \> \> $Q.cost := Q.cost + (-\log a_{iN})$\\ $\Box$
\end{tabbing}
At time $T$ the token in the exit state of the LD can be back-tracked
and the most likely word sequence can be established. Note that OD
HMMs are never deactivated in the algorithm above. In the actual
implementation however the Beam-Search heuristic is used which means
that OD HMMs are deactivated if their costs exceed a threshold. The
threshold is continuously updated according to the best scoring OD
HMM.

\section{Experiments}
\label{sec:exp}
{\sc linlin} \cite{Jonsson:95} is a natural language dialogue system
that takes queries in Swedish as input and produces SQL-queries that
can be fed to a DBMS. {\sc linlin} can be hooked up with a couple of
different databases. The utterances below are from a corpus collected
with {\sc linlin} connected to a database with information on used
cars. The corpus is called {\sc cars} and was collected with the
`wizard of Oz' method, cf. Dahlb\"{a}ck \cite{DJA:93}. {\sc cars}
includes 20 dialogues with a total of 369 user utterances. In 71 of
these there is one or more lexical error. This makes approximately one
in every five user utterances erroneous only with respect to lexical
errors. There is a total of 95 lexical errors of which 60 are
misspellings, 20 run-ons and 15 splits.The lexical error categories
are misspellings and segmentation errors. The segmentation errors can
be further divided into run-ons and splits. The user
utterances\footnote{The utterances in this section are word-for-word
Swedish to English translations where the crucial aspect of an
utterance has been preserved. Hyphens that do not wrap a line indicate
Swedish noun compounds.}  (\ref{eq:miss}) through (\ref{eq:split})
from {\sc cars} show the three basic lexical error categories:
misspellings, run-ons and splits respectively.

\begin{list}{{\tt ==>}}{\setlength{\rightmargin}{2cm}}
        \inputitem{equation}What is the maintenance-cost for the
        respective models in the abo\error{u}e table?\label{eq:miss}
        \inputitem{equation}Same question but at most\error{1}4
        s\label{eq:ro} \inputitem{equation}only those with
        coup\'{e}\error{\ }space 3-4\label{eq:split}
\end{list}

\noindent The tiny stars indicate where a lexical error has occurred ({\tt
  coup\'{e} space} should be {\tt coup\'{e}-space}).

To test our ideas with the layered HMM approach in the token passing
framework we have developed a system {\sf ctr} to perform connected
text recognition. The {\sf ctr} experiments reported here concern the
{\sc cars} corpus. The intention of these experiments is of course to
get an indication of the performance of the technique presented in the
preceeding section, and to see whether a linguistic model will help at
all in the recovery from lexical errors. In general this is obviously
true but with sparse data it is not so certain. We are also interested
to see what impact word classification along different linguistic
dimensions will have on the performance of {\sf ctr}. We have tried a
relatively rich syntactic class-set and a smaller domain oriented
class-set.

\begin{figure}[b]
   \centerline{\psfig{figure=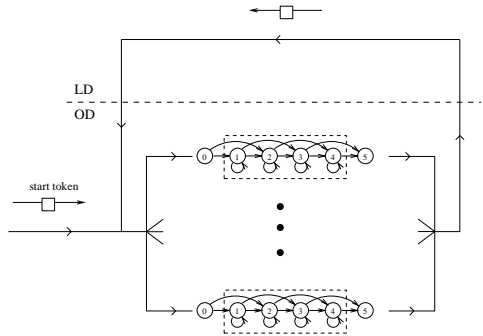}}
   \caption{The Baseline experiment -- {\sf ctr} setup\label{fig:ldod-naive}}
\end{figure}

Experiments have been conducted on three different language models, a
unigram language model and two biclass language models. We also have a
baseline to which the results of these experiments can be
compared. The baseline experiment involved no linguistic constraints
so the correction of lexical errors was done by the orthographic
decoder alone. See Figure~\ref{fig:ldod-naive} for the baseline {\sf
ctr} setup.

The 20 dialogues were randomly divided into five parts of four
dialogues each. In the experiments, 16 dialogues (four parts) were
used to obtain the language model and then the model was tested on the
remaining four dialogues (one part). The partitionings were rotated so
that each language model was tested on all of the five parts. The same
orthographic decoder was used in all the experiments.

\subsection{The Unigram Language Model}
\label{subsec:unigram}
The unigram language model:
\[\prob{w_{1},\ldots ,w_{T}} = \prod_{i=1}^{T}\prob{w_{i}}\]
The language model's parameters are extracted from the training corpus
of the five partitionings.
\[\prob{w_{i}} = \frac{Count(w_{i})}{N}\]
Where $N$ is the number of word tokens in the training corpus. The
words that did not show up in the training corpus was smoothed with
the simple smoothing scheme described by Levinson {\em et al.}
\cite{levinson+al:83} (p. 1053), sometimes referred to as additive
smoothing. A small probability mass is reserved for unseen events.

The linguistic decoder realizing the unigram model is a single state
HMM (three states including the entry and exit states). The parameters
of the unigram make up the observation symbol distribution of the LD
HMM.

\subsection{The biclass-dom Language Model}
\label{subsec:biclassdom}
In the biclass-dom language model there are 19 word classes. The words
of the corpus are grouped into classes that are semantically, or,
domain oriented, thus the name biclass-dom. Examples of classes and
class members are: OH -- Object Head -- (\mbox{`{\tt saab 900}'},
\mbox{`{\tt all}'}\ldots), AH -- Aspect Head -- (\mbox{`{\tt costs}'},
\mbox{`{\tt acceleration}'}\ldots), CH -- Communicative Head --
(\mbox{`{\tt show}'}, \mbox{`{\tt example}'}\ldots). If $Cl_{1}^{T+1}$
denotes a sequence of $T$ classes assigned to a sequence of $T$ words,
(plus the dummy class $Cl_{T+1}$ corresponding to the nonexistent word
$w_{T+1}$), the biclass language model looks like:

\[\prob{w_{1},\ldots ,w_{T}} = \sum_{all
Cl_{1}^{T+1}}\prod_{i=1}^{T}\prob{w_{i}|Cl_{i}}\prob{Cl_{i+1}|Cl_{i}}\]

\begin{figure}[ht]
   \centerline{\psfig{figure=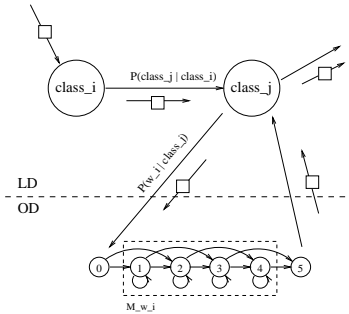}}
   \caption{The biclass experiments -- {\sf ctr} setup\label{fig:ldod-interact}}
\end{figure}

See Figure~\ref{fig:ldod-interact} for the {\sf ctr} setup. The
language model's parameters are extracted from the tagged training
corpus of the five partitionings:

\[\prob{Cl_{i+1}|Cl_{i}} = \frac{Count(Cl_{i},Cl_{i+1})}{Count(Cl_{i})}\]
\[\prob{w_{i}|Cl_{i}} = \frac{Count(Cl_{i},w_{i})}{Count(Cl_{i})}\]

In the biclass-dom case both the state transition distribution and the
observation symbol distribution have to be smoothed. It is done with
additive smoothing.

\subsection{The biclass-SUC Language Model}
\label{subsec:biclasssuc}
The biclass-SUC language model has 31 classes. The set of classes
originates from the SUC corpus (Stockholm-Ume\aa\ Corpus
\cite{kallgren:col:90}). The classes used in the SUC corpus are
traditional part-of-speech with associated morphological features. We
have made slight modifications to the original set of SUC-tags to
obtain a set of atomic classes with (supposedly) different syntactic
distributions. Examples of classes and class members are: PM -- Proper
noun -- (`{\tt saab 900}',\ldots), DT -- Determiner -- (`{\tt
all}',\ldots), VBF -- Verb form finite -- (`{\tt costs}',\ldots), NN
-- Noun -- (`{\tt acceleration}', `{\tt example}',\ldots), VBP -- Verb
form imperative -- (`{\tt show}',\ldots).

The biclass-SUC parameters are extracted from the tagged training
corpus in the same way as was done with biclass-dom. Also with
biclass-SUC both the state transition distribution and the observation
symbol distribution are smoothed with the additive smoothing scheme.

\subsection{The Orthographic Decoder (OD)}
\label{subsec:od}
The OD contains 584 word modeling HMMs, one for each word type in the
corpus. The structure of the OD HMMs can be seen in
Figure~\ref{fig:m-show-space}. Ideally each HHM should be trained on
typical errors occurring in Swedish text. Unfortunately there is no
such error corpus available and we can certainly not train the HMMs on
the errors occurring in the corpus. We must find a way to generate an
error corpus so that the OD HMMs can be trained and used for other
purposes as well, not just to identify the particular errors in this
corpus.

The primitive error types on the character level are: deletion
(e.g. `{\tt\ shw}'), insertion (e.g. `{\tt\ shiow}'), substitution
(e.g. `{\tt\ shiw}') and transposition (e.g. `{\tt\ sohw}'). Although
these error types apply to the space character as well as to any other
character, we have an extra error type dealing only with the space
character, it is called white space insertion (e.g. `{\tt\ sh
ow}'). There is also an error type called double stroke (e.g. `{\tt\
shoow}'). The error types insertion and substitution raises the
question what to insert and what to substitute for respectively. One
hypothesis is that keyboard neighbours are likely to take part in
e.g. substitutions. The neighbours\footnote{The neighbour relation is
limited to left and right neighbours.} to `{\tt o}' are `{\tt i}' and
`{\tt p}', so if substitutions are applied, the error corpus for
${\cal M}_{show}$ will contain amongst others `{\tt\ shiw}' and `{\tt\
shpw}'. If insertions are applied it will also contain `{\tt\ shiow}',
`{\tt\ shoiw}', `{\tt\ shpow}' and `{\tt\ shopw}'.

In these experiments the error corpora were generated with the error
types substitution, deletion and white space insertion. When an error
corpus is generated the selected error types are applied to each
character position in the word modeled by the trainee. Apart from this
general strategy, some special words need specialized corpora. These
words include single character `words' such as `{\tt\ =}', `{\tt\ ?}',
`{\tt\ .}' and so on. The relatively few numbers occurring in the
corpus also have their own HMM. These special words have corpora
generated with only the white space insertion error type. The OD HMMs
were trained with the Baum-Welch reestimation algorithm. After
training each HMM had their observation symbol distribution smoothed
with the additive method.

Note that we are evading the unknown word problem. Even if a word type
is unseen in the training corpus of an experiment, the OD will still
contain the model corresponding to the unseen word.

\subsection{Results}
\label{subsec:res}
When an experiment is conducted, {\sf ctr} is run on the corpus in
batch mode, i.e. utterances are processed from an input file and
output to an output file. This creates pairs of utterances. Resulting
from an experiment is thus a set of pairs \mbox{$\langle${\tt original
utterance}\ ,\ {\tt normalized utterance}$\rangle$}. An experiment is
evaluated by comparing the pairs resulting from the experiment to
pairs in a result {\em key}. The key is a hand-made set of pairs where
the first element (the original utterance) contains at least one
lexical error and the second element is the appropriate correction of
that utterance. This set is called $\mbox{\boldmath $A$}$. The {\em
outcome} of an experiment are the pairs produced in the experiment
where the second element is {\em not} identical to the first one, or,
the first element is identical to the first element in one of the
pairs in $\mbox{\boldmath $A$}$. This set is called $\mbox{\boldmath
$C$}$. The pairs in the outcome that are also in the key belong to the
set $\mbox{\boldmath $B$}$, i.e. $\mbox{\boldmath $B$}=\mbox{\boldmath
$A$}\cap\mbox{\boldmath $C$}$. The outcome of an experiment can now be
rated with respect to the performance measures {\em recall} and {\em
precision}.

\begin{eqnarray*}
recall & = & \frac{\mid\mbox{\boldmath $B$}\mid}{\mid\mbox{\boldmath
$A$}\mid}\times 100\\\\ precision & = & \frac{\mid\mbox{\boldmath
$B$}\mid}{\mid\mbox{\boldmath $C$}\mid}\times 100
\end{eqnarray*}

An example of a pair in $\mbox{\boldmath $A$}$: $\langle${\tt rust
prote\error{t}ion for\error{t}hese}\ ,\ {\tt rust protection for
these}$\rangle$. The first element of the pair contains two errors and
we like to extend the performance measure to account for individual
errors, not just whole utterances. From the outcome of the experiment
we can extract the counterparts for $\mbox{\boldmath $A$}$,
$\mbox{\boldmath $B$}$ and $\mbox{\boldmath $C$}$ that applies to the
respective error categories. We have $\mbox{\boldmath $A^{m}$}$,
$\mbox{\boldmath $B^{m}$}$ and $\mbox{\boldmath $C^{m}$}$ for
misspellings, $\mbox{\boldmath $A^{r}$}$, $\mbox{\boldmath $B^{r}$}$
and $\mbox{\boldmath $C^{r}$}$ for run-ons and we have
$\mbox{\boldmath $A^{s}$}$, $\mbox{\boldmath $B^{s}$}$ and
$\mbox{\boldmath $C^{s}$}$ for splits. We are also interested in the
total number of individual errors so the key $\mbox{\boldmath
$A^{tot}$}=\mbox{\boldmath $A^{m}$}\cup\mbox{\boldmath
$A^{r}$}\cup\mbox{\boldmath $A^{s}$}$ is added to the list of
keys. The example pair above that was a member of $\mbox{\boldmath
$A$}$ also adds \mbox{$\langle${\tt prote\error{t}ion}\ ,\ {\tt
protection}$\rangle$} to $\mbox{\boldmath $A^{m}$}$ and
$\mbox{\boldmath $A^{tot}$}$ and \mbox{$\langle${\tt
for\error{t}hese}\ ,\ {\tt for these}$\rangle$} adds to
$\mbox{\boldmath $A^{r}$}$ and $\mbox{\boldmath $A^{tot}$}$. The five
keys provide the five performance categories in the tables below. The
tables below show the joint results for the disjunct test corpora of
the five partitionings. This means that each language model has been
tested on the entire corpus, only it has been done in five steps with
five different training corpora.

\begin{table}[htb]
\centerline{
\begin{tabular}{||c|l|c|c||}
\hline Experiment & Performance categories & Recall & Precision \\
\cline{1-4} & utterances & 73 \% & 73 \% \\ \cline{2-4} & total & 80
\% & 77 \% \\ \cline{2-4} Baseline & misspellings & 74 \% & 75 \% \\
\cline{2-4} & run-ons & 100 \% & 100 \% \\ \cline{2-4} & splits & 79
\% & 58 \% \\ \cline{2-4} \hline
\end{tabular}
}

\caption{Baseline experiment\label{tab:bas}}
\end{table}

In the baseline experiment (Table~\ref{tab:bas}) there is an 80\%
total recall. The drop in precision is quite small which is not
surprising since there is no language model to `disturb' the
orthographic decoder. The 80\% $\rightarrow$ 77\% drop is altogether
due to the bad splits precision. In a handfull of places in the corpus
there are double space characters in between words. Since the LD does
not add a cost to the forming of words, the superfluous space will be
changed to a single character word such as {\tt `,'}. The double space
in the input utterance does not constitute an error by our definition,
so an error is introduced and the error is classified in terms of the
transformation from input to output utterance. For example: {\tt
`\ldots models\symbol{32}\symbol{32}and\ldots'} is transformed into
{\tt `\ldots models,\symbol{32}and\ldots'}.

When the LD is furnished with the unigram language model
(Table~\ref{tab:uni}) performance is enhanced on all categories. The
total enhancement (80\% $\rightarrow$ 86\%) compared to the baseline
is due to improved ability to deal with misspellings and splits. On
four accounts the unigram model was able to make the right decision on
`close calls' regarding misspellings that the baseline failed to deal
with.

\begin{table}[htb]
\centerline{
\begin{tabular}{||c|l|c|c||}
\hline Experiment & Performance categories & Recall & Precision \\
\cline{1-4} & utterances & 83 \% & 76 \% \\ \cline{2-4} & total & 86
\% & 80 \% \\ \cline{2-4} Unigram & misspellings & 81 \% & 76 \% \\
\cline{2-4} & run-ons & 100 \% & 83 \% \\ \cline{2-4} & splits & 93 \%
& 93 \% \\ \cline{2-4} \hline
\end{tabular}
}

\caption{Experiments with the Unigram language model\label{tab:uni}}
\end{table}

\begin{table}[htb]
\centerline{
\begin{tabular}{||c|l|c|c||}
\hline Experiment & Performance categories & Recall & Precision \\
\cline{1-4} & utterances & 87 \% & 79 \% \\ \cline{2-4} & total & 93
\% & 83 \% \\ \cline{2-4} Biclass-dom & misspellings & 89 \% & 79 \%
\\ \cline{2-4} & run-ons & 100 \% & 87 \% \\ \cline{2-4} & splits &
100 \% & 100 \% \\ \cline{2-4} \hline
\end{tabular}
}

\caption{Experiments with the biclass-dom language model\label{tab:bidom}}
\end{table}

Both the biclass experiments
(Tables~\ref{tab:bidom}~\&~\ref{tab:bisuc}) show steady improvement
over both the baseline and the unigram. Mutually however, between the
biclass-dom and the biclass-SUC, there is not much
difference. Biclass-SUC seems to have a narrow advantage with respect
to precision, but the two biclass language models exhibit virtually
the same results. The advantage that biclass-SUC has because of the
richer class-set is possibly neutralized by the poorer estimates
resulting from the added data sparseness problem. If the result that
domain classes yield as good performance as syntactic classes would
extrapolate to a bigger corpus, we would consider this a positive
result in the context of a dialogue system since the interpretation
step (input query $\rightarrow$ SQL-query) is substantially reduced by
the domain-classification of input words.

\begin{table}[htb]
\centerline{
\begin{tabular}{||c|l|c|c||}
\hline Experiment & Performance categories & Recall & Precision \\
\cline{1-4} & utterances & 91 \% & 84 \% \\ \cline{2-4} & total & 93
\% & 85 \% \\ \cline{2-4} Biclass-SUC & misspellings & 89 \% & 83 \%
\\ \cline{2-4} & run-ons & 100 \% & 80 \% \\ \cline{2-4} & splits &
100 \% & 100 \% \\ \cline{2-4} \hline
\end{tabular}
}

\caption{Experiments with the biclass-SUC language model\label{tab:bisuc}}
\end{table}

All the experiments show a decline in performance from recall to
precision. The reason is of course that (almost) all lexical errors in
the test corpus are `detected', i.e. character sequences not in the
vocabulary will be changed. In the corpus there is one case where an
accidental misspelling turns out to be a different legal lexical
construction. The effect can be seen in Table~\ref{tab:bas}. This is
the only way that a lexical error can go unnoticed and precision be
higher than recall. There are also four real-word error splits (one of
the tokens is a real word), which can all be handled by the two
biclass models.

\section{Concluding Remarks}
\label{sec:rem}
Results indicate that the {\sf ctr} system can be used for removing
many of the lexical errors in the input to a natural language
interface like {\sc linlin}. The ratio of utterances that are affected
by lexical errors is brought down from \(\sim 20\%\ (71/369)\
\mbox{to} \sim 3\%\ (12/369)\) in the biclass-SUC experiment in
Table~\ref{tab:bisuc}. It is not easy to compare the results presented
here to those of Carter~\cite{carter:92}, but keeping in mind that
Carter exploits the full-fledged syntactic and semantic capabilities
of {\sc clare}, these figures compare quite favourably, albeit Carter
uses a more realistically sized lexicon (1600 root forms).

The correctness criterion for the repairs suggested by {\sf ctr} is
quite harsh. String equality is the measure used and some of the bad
repairs could probably be handled by a parser. There are examples of
adjective-noun sequences in the input that have been run together to
form noun compounds which do not change the meaning of the utterance
(much). These show up in the precision performance for run-ons. {\sc
cars} also contains some `impossible' lexical errors. Examples of
which are: the single character utterance {\tt `s'}\footnote{{\sf ctr}
suggested {\tt `so'} as a repair, but we had decided that the subject
probably meant {\tt `show'}} and two `new' abbreviations, {\tt `ins'}
for {\tt `instead'} (two instances) and {\tt `value-decs'} for {\tt
`value-decrease'}. If it were not for these four errors the recall
performance for the biclass models would be 97\%.

{\sc cars} is a small corpus. Even with the relatively weak language
models used in the experiments, the data sparseness is evident. The
data sparseness is emphasized by the fact that precision is overall
worse than recall in spite of the weak models. With the baseline no
linguistic disturbance is introduced, while particularly biclass-dom
has a relatively poor precision. But still, the net return of the
models is positive in all three cases.

The calculations in the Token Passing algorithm are performed
incrementally, character by character, as the user enters an input
utterance. This means that lexical error recovery can be performed on
the fly, without the user knowing about it.

The approach presented here can of course be used in applications
other than dialogue systems, although it is unlikely that it will be
practical for unrestricted text. The size of the vocabulary, the
number of OD HMMs, will be too large. What happens to the performance
of {\sf ctr} when the vocabulary is extended, is one of the questions
that future experiments will have to answer.



\end{document}